\documentclass{bmvc2k}

\graphicspath{{./Figures/}}
\usepackage[font={color=bmv@captioncolor}]{caption}
\usepackage[font={color=black}]{subcaption}
\newcommand*\phantomsubfigure[1]{\begin{subfigure}[]{0pt}\phantomcaption\label{#1}\end{subfigure}}
\usepackage{multicol}
\usepackage{afterpage}

\title{\textit{Hidden-Fold Networks}: Random Recurrent Residuals Using Sparse Supermasks}

\addauthor{Ángel López García-Arias}{lopez@artic.iir.titech.ac.jp}{1}
\addauthor{Masanori Hashimoto}{hashimoto@i.kyoto-u.ac.jp}{2}
\addauthor{Masato Motomura}{motomura@artic.iir.titech.ac.jp}{1}
\addauthor{Jaehoon Yu}{yu.jaehoon@artic.iir.titech.ac.jp}{1}

\addinstitution{AI Computing Research Unit, \\ Tokyo Institute of Technology, Japan}
\addinstitution{Department of Communications \\and Computer Engineering, \\ Kyoto University, Japan}

\runninghead{López, Hashimoto, Motomura, Yu}{Hidden-Fold Networks}

\begin{document}

\maketitle

    \vspace{-0.35cm}
\begin{abstract}
    Deep neural networks (DNNs) are so over-parametrized that recent research has found them to already contain a subnetwork with high accuracy at their randomly initialized state. Finding these subnetworks is a viable alternative training method to weight learning. In parallel, another line of work has hypothesized that deep residual networks (ResNets) are trying to approximate the behaviour of shallow recurrent neural networks (RNNs) and has proposed a way for compressing them into recurrent models. This paper proposes blending these lines of research into a highly compressed yet accurate model: Hidden-Fold Networks (HFNs). By first folding ResNet into a recurrent structure and then searching for an accurate subnetwork hidden within the randomly initialized model, a high-performing yet tiny HFN is obtained without ever updating the weights. As a result, HFN achieves equivalent performance to ResNet50 on CIFAR100 while occupying 38.5x less memory, and similar performance to ResNet34 on ImageNet with a memory size 26.8x smaller. The HFN will become even more attractive by minimizing data transfers while staying accurate when it runs on highly-quantized and randomly-weighted DNN inference accelerators.
    Code available at: \url{https://github.com/Lopez-Angel/hidden-fold-networks}
    
\end{abstract}
\vspace{-0.15cm}
    \vspace{-0.25cm}
\section{Introduction} \label{sec:introduction}
\vspace{-0.1cm}

    Deep neural networks (DNNs) have followed a trend of improvement in accuracy by growing in size. Current popular models, such as residual networks (ResNets)~\cite{resnet}, have tens of millions of parameters. 
    Since off-chip memory access uses much more energy and time than arithmetic computation, data transfers dominate the high cost of DNN processing. Various lines of research have found that these DNNs are much larger than they need to be and have proposed methods for compressing them without harming their accuracy. 
    
    Several initiatives have attempted to move the weight of learning away from learning weights. Perturbative neural networks~\cite{perturbative_neural_networks} proposed substituting convolutional layers with perturbative layers: layers that add fixed random noise 
    to the inputs. Although with lower accuracy, these fixed random modifications together with simple $1$x$1$ convolutional layers are claimed to be enough for learning.
    A similar approach in this direction showed that learning the parameters of the batch normalization layers while keeping the weights untouched is enough to train a neural network~\cite{training_on_batchnorm}.

	Meanwhile, network pruning efforts have made it evident that most popular models are over-parametrized and that large parts of them can be pruned without affecting accuracy~\cite{songhan_weightsandconnections}, leading to highly compressed models~\cite{deepcomp} and efficient specialized hardware~\cite{han2016eie}. 
	It was generally found that the subnetworks resulting from pruning were hard to train from scratch. However, the Lottery Ticket Hypothesis~\cite{lottery_ticket} showed that over-parametrized DNNs contain a subnetwork---referred to as a lottery ticket---that can be trained in isolation, overperforming the whole network while being smaller, and requiring fewer iterations to learn. Its authors also proposed a way of finding these subnetworks through the iterative application of three steps: training, pruning, and re-initialization.

    This idea was taken a step further with the discovery of hidden-networks (HNNs)~\cite{supermask,hidden-networks}: inside a randomly initialized DNN, there is a hidden subnetwork that, without being trained, achieves similar accuracy to the whole network trained. HNNs can be found by modifying the learning algorithm to optimize a binary mask of the weights---a supermask.
    Reference~\cite{hidden-networks} proposes an algorithm that finds in ResNet a subnetwork with an accuracy similar to that obtained by training the whole model with dense learned weights. 
	Moreover, a single network can be used for multiple tasks without catastrophic forgetting by finding an appropriate mask for each task or even using combinations of masks for new tasks~\cite{supsup}.
    
    A different method of compressing networks is found in the hypothesis that ResNets may be approximating unrolled shallow recursive neural networks (RNNs), and that the gains from additional layers correspond to additional recursive iterations~\cite{liao_poggio,folding_schmidhuber}. This theory is supported by the authors of~\cite{folding_bengio}, who found that ResNet's deeper residual blocks are learning to perform iterative refinement of features. Moreover, the work in~\cite{liao_poggio} demonstrated that folding ResNet into a 4-layer RNN only had a moderate impact on accuracy. 
    
    This theory has roots in neuroscientific observations that have compared the DNNs used for image recognition with biological visual systems. Although DNNs are excellent models of the primate visual cortex~\cite{yamins2014performance,cichy2016deep}, there are some essential differences. One striking difference is observed at the architectural level: where DNNs typically have tens or hundreds of layers, the visual ventral stream in primate brains has just between four and six layers~\cite{cornet}. Furthermore, the visual ventral stream layers function in a recurrent way via multiple types of lateral connections~\cite{kar2019evidence}, unlike popular feed-forward DNN. 
    The hypothesis conjectured on this evidence is that deep learning models are converging to structures similar to the brain's visual system, and that this convergence can be accelerated by drawing inspiration from these differences. Following these ideas, shallow recurrent models have resulted in both more efficient~\cite{iamnn} and more brain-like~\cite{cornet} networks.

    This paper blends these two research trends into a new type of network: Hidden-Fold Networks (HFNs). First, ResNets are transformed into recursive models through the use of shared weights. Inside this randomly initialized folded structure lies hidden a high-performing network---an HFN. HFNs are unearthed with supermask training. Due to its tiny number of parameters and memory requirements, the proposed method is an exceptional candidate for implementing energy-efficient DNN acceleration hardware. 
    \vspace{-0.4cm}
\section{Proposed method: Hidden-Fold Networks} \label{sec:Methods}
\vspace{-0.1cm}
     There are three main intuitions for combining into HFN the research trends of learning without updating weights and folding, and for the synergy between their respective techniques.
    \begin{itemize}
        \vspace{-0.2cm}\item 
        Both trends recognize that modern DNNs are over-parametrized and capitalize on it in a way that provides model compression with a minor loss of accuracy. 
        \vspace{-0.2cm}\item
        Supermask training modifies how weights are treated, whereas folding changes the architecture. The techniques are orthogonal, and therefore it should be possible to combine them without interference. 
        \vspace{-0.2cm}\item 
        ResNet becomes equivalent to a recurrent model when blocks with the same shape are forced to use shared weights. Similarly, the weight initialization used for HNN initializes blocks of the same shape to values of the same modulus. Therefore, both techniques make ResNet more similar to an RNN.
    \end{itemize}\vspace{-0.15cm}

    To find an HFN, first a ResNet is initialized randomly. Then, blocks of identical shape within each stage are transformed into recurrent blocks by sharing their weights, except for the BatchNorm parameters. Lastly, instead of updating the weights through backpropagation, the HFN is searched for by training a supermask.
    The proposed method is formed by four components: architecture, weight initialization, training method, and batch normalization. We describe their details in order.
        
    \textbf{Architecture --- Folded ResNet:}\quad
    This work uses ResNets~\cite{resnet}, converting them into recurrent models via folding~\cite{liao_poggio}.
    ResNets are formed by an input convolutional layer, a main network divided in four stages, and a fully connected output layer, as depicted in \figurename~\ref{foldin_n_arch_a}. Stages are composed of multiple bottleneck blocks, each of which is formed by three convolutional layers. ResNets are named after their total number of convolutional layers, including input and output layers. 
    Since each stage is formed by chaining blocks of the same shape, if their weights are shared, then the function they perform becomes identical. Chaining identical functions is equivalent to applying a function iteratively. Therefore, blocks at each stage can be converted into a recurrent block by sharing their weights, as depicted in \figurename~\ref{foldin_n_arch_b}. This process is the reverse of RNN unrolling and is referred to as folding~\cite{liao_poggio}.
    
	\begin{figure}[t]
        \centering
        \includegraphics[width=\linewidth]{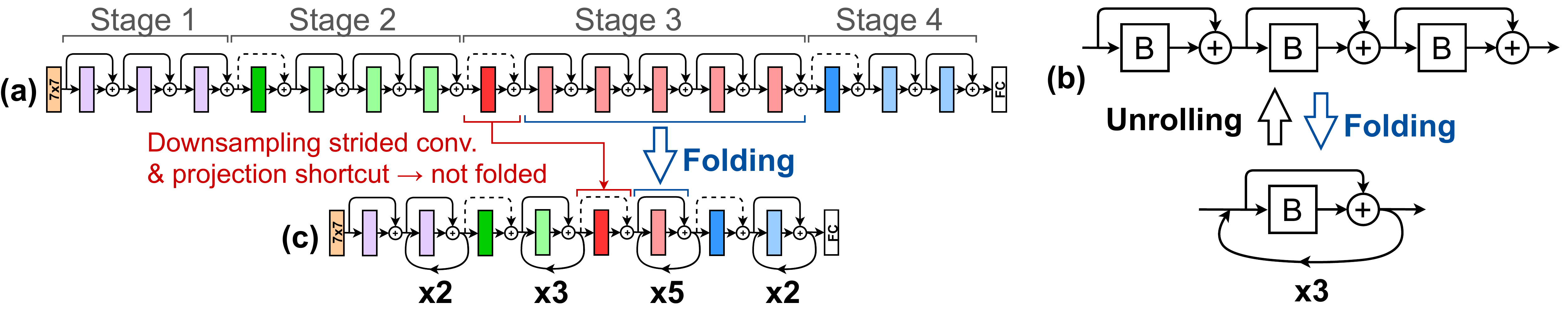}
        \phantomsubfigure{foldin_n_arch_a}
        \phantomsubfigure{foldin_n_arch_b}
        \phantomsubfigure{foldin_n_arch_c}
        \vspace{0.25cm}
        \caption[caption]{Architecture of HFN
        (a) ResNet50. ResNets are formed by an input layer, four main stages, and an output layer. 
        (b) Repeated identical instances of a structure are equivalent to a recurrent structure.
        (c) In this example, identically shaped bottleneck blocks in all stages of ResNet50 are folded.}
        \label{fig:foldin_n_arch}
        \vspace{-0.6cm}   
    \end{figure} 
    
    The difference between stages is the size of the feature maps. To adjust for these dimensional differences, the first block of each stage uses downsampling strided convolution and projection shortcuts. Since this first block is different from the rest, it cannot be folded. These projection blocks could be eliminated, as explored in~\cite{liao_poggio} and~\cite{folding_bengio}, but following the intuition that these dimensional transformations correspond to compositional changes in the level of representation~\cite{folding_schmidhuber}, this paper keeps them. The rest of the blocks are folded into a single recurrent block, which is iterated a number of times equal to the number of blocks folded (see \figurename~\ref{foldin_n_arch_c}). Since folding makes sense only with more than two blocks per stage, the smallest possible HFNs are ResNet34 and ResNet50. This paper only uses ResNets with bottleneck blocks, and therefore the smallest model used in this paper is ResNet50. The wide variants of ResNet~\cite{wide_resnets} are also used to consider the depth-width tradeoff.
    
\begin{figure}
        \centering
        \begin{subfigure}[b]{0.51\textwidth}
            \centering
            \includegraphics[width=\textwidth]{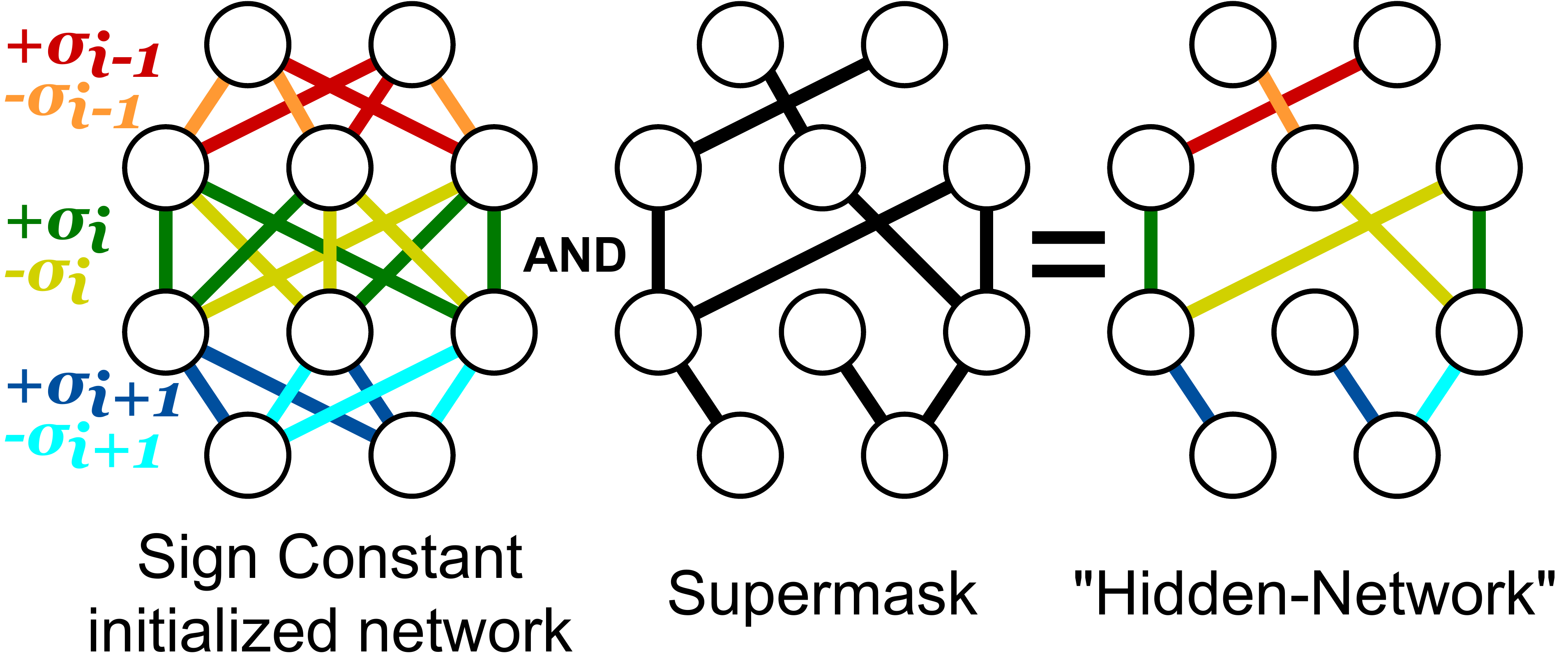}
            \caption{Supermask application}
            \label{fig:hidden_networks}
        \end{subfigure}\hfill
        \begin{subfigure}[b]{0.46\textwidth}
		  \centering
		  \includegraphics[width=\linewidth]{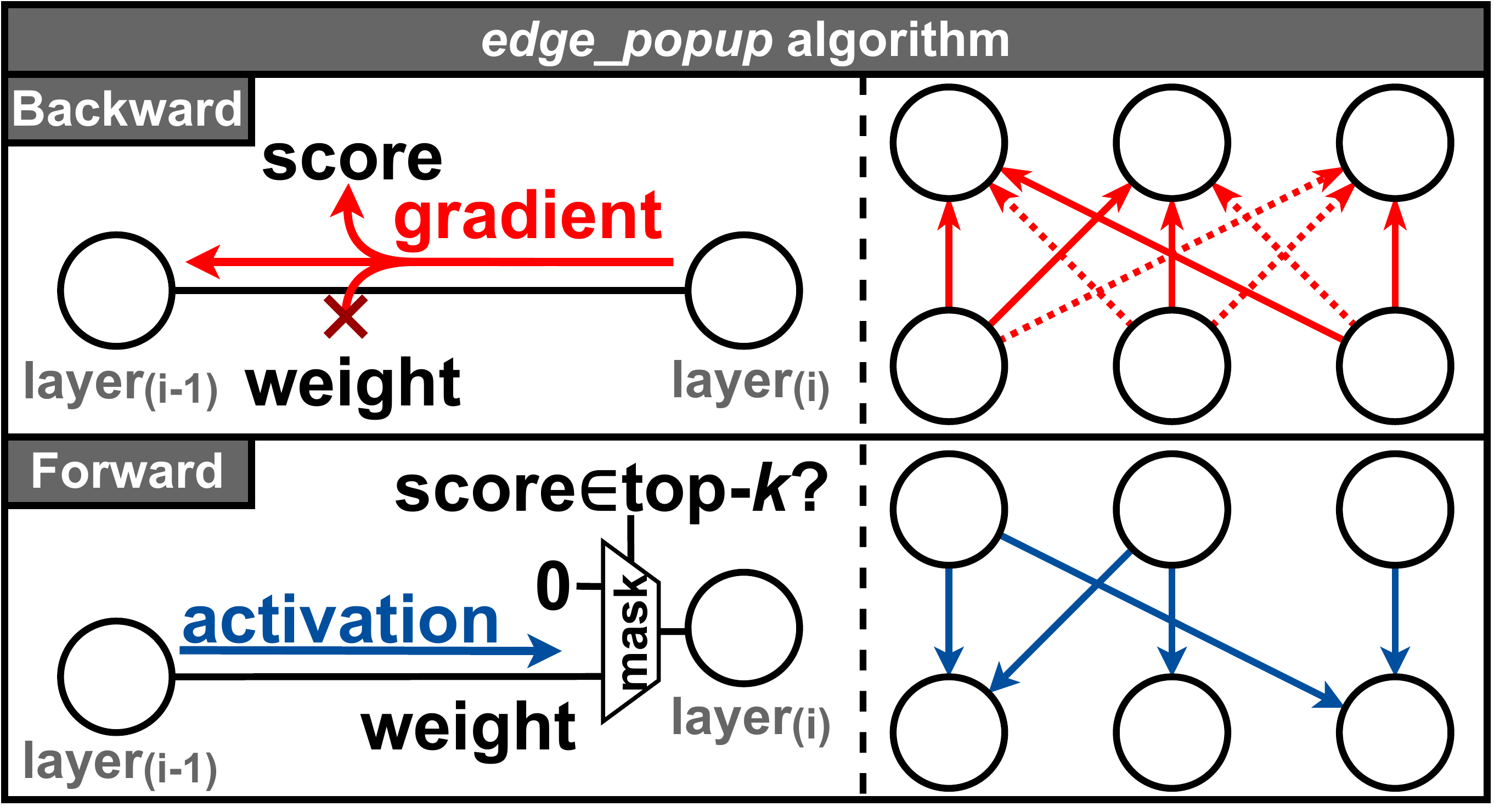}
		  \caption[caption]{The \texttt{edge\_popup} algorithm~\cite{hidden-networks}}
		  \label{fig:edge_popup}
		\end{subfigure}
        \vspace{0.25cm}
        \caption{(a) The figure shows a supermask applied during inference to a network initialized with SC. Each colour represents each layer's $\sigma_i$ value, with two hues for sign ($-\sigma_i$ and $+\sigma_i$).
        (b) Instead of the weights, backpropagation updates the score assigned to each connection. The connections with the top-$k$\% scores (solid lines) of each layer form the supermask.}
        \label{fig:supermasks_n_sc}
        \vspace{-0.6cm}   
    \end{figure}

    \textbf{Weight initialization --- Signed Kaiming Constant (SC):}\quad
    Weight initialization is done as~\cite{hidden-networks}. The network initialization process is crucial to find high-performance subnetworks. Good results are achieved using Kaiming initialization~\cite{kaiming_init}, which determines the weight's distribution by taking into account the network's shape. Remarkably, even better results are obtained with SC initialization, which initializes all the weights of a layer to the standard deviation $\sigma$ of Kaiming normal distribution, and sets each weight's sign randomly. In other words, the weights are initialized by sampling uniformly from $\{-\sigma, \sigma\}$, with an independent $\sigma$ for each layer (see \figurename~\ref{fig:hidden_networks}).
    
    \begin{figure}[b]
        \vspace{-0.6cm} 
        \centering
        \includegraphics[width=\linewidth]{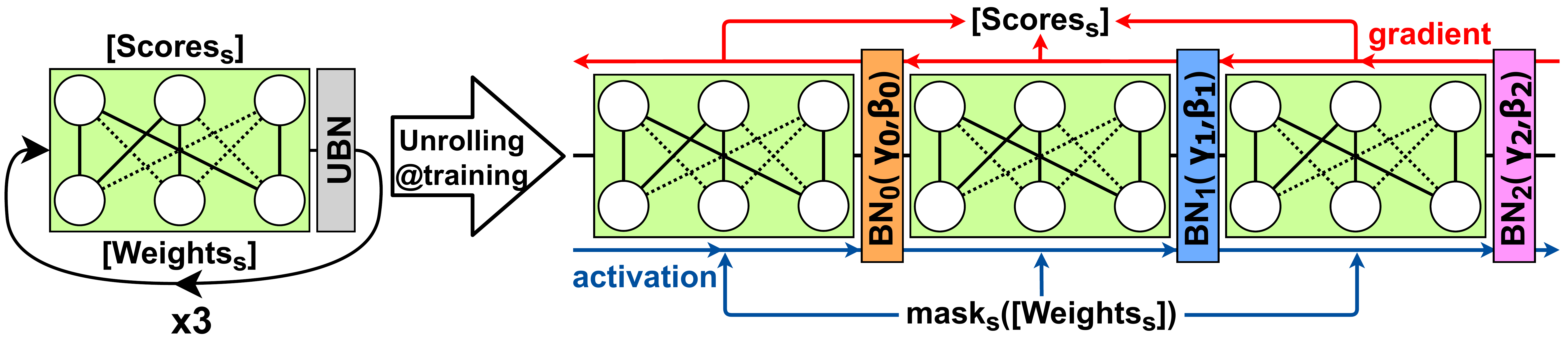}
        \vspace{0.25cm}
        \caption[caption]{How a folded block is unrolled during training time. In all iterations the same mask and weights are used, and gradients update the same scores. However, an independent affine BatchNorm layer is learned for each iteration.}
        \label{fig:unrolled_hfn}
    \end{figure}      
    
    \textbf{Training method --- Supermasks and \texttt{edge\_popup}:}\quad
	An HFN is found by optimizing a supermask with the \texttt{edge\_popup} algorithm~\cite{hidden-networks}, summarized in \figurename~\ref{fig:edge_popup}. Each connection is assigned a random score in addition to a weight. In the backwards pass, backpropagation updates the scores instead of the weights, which are left in their randomly initialized state. 
	The top-$k$\% highest scores of each convolutional layer are selected in the forward pass, and only their respective connections are used. This is done by constructing a binary mask of the weights---a supermask---in which the positions corresponding to the connections with the top-$k$\% highest scores are set to $1$, while the rest are deactivated by assigning them a $0$. As shown in \figurename~\ref{fig:hidden_networks}, the supermask is applied to the weight tensors through an element-wise multiplication (or a logic \texttt{AND} function) during the forward pass, but not during the backward pass, when gradients are propagated to all scores. Thus, top-$k$\% is a measure of the model's density. Scores are unnecessary for inference, but since supermasks are task and weight-dependent, the same random seed must be used for training and inference. This method is used for all convolutional layers of HFN. Folded blocks share a common supermask, which receives gradients from each iteration (see \figurename~\ref{fig:unrolled_hfn}).
    
    \textbf{Batch Normalization~\cite{batchnorm}:}\quad
    Following~\cite{hidden-networks}, supermask training uses non-affine BatchNorm, i.e., BatchNorm whose learnable parameters are fixed (bias $\beta = 0$ and scale $\gamma = 1$). However, folded ResNets suffer from overfitting and exploding layer activations. Reference~\cite{folding_bengio} suggested using unshared batch normalization (UBN) to alleviate this problem. UBN consists of not sharing batch normalization's learnable parameters ($\beta$ and $\gamma$) between folded blocks, having a different set for each iteration instead, as exemplified in \figurename~\ref{fig:unrolled_hfn}. This approach also proved successful in~\cite{cornet}.
    All BatchNorm layers used are non-affine, except for folded blocks, which use UBN.
    \vspace{-0.4cm}
\section{Experiments} \label{sec:Experiments}
\vspace{-0.1cm}
    This section compares the four methods summarized in \tableautorefname~\ref{table:methods}. The experiments described were developed using as a base the PyTorch~\cite{pytorch} code released by the authors of~\cite{hidden-networks}, which can be found at~\cite{hidden-networks-github}.
    After confirming the compatibility of supermasks and folding, the optimal HFN configuration is determined by comparing on the same ResNet50. Then, the obtained configuration is used to compare ResNets of different depths and widths. Lastly, the memory advantages of HFN are estimated considering compression on specialized hardware.

    Experiments were carried out on the CIFAR100~\cite{CIFAR} and ImageNet~\cite{Imagenet} datasets. Since the smallest model used for HFN is the rather big ResNet50, smaller datasets were not used.
    Unless explicitly stated otherwise, experiments were carried using the following configurations. In experiments using CIFAR100, the $60,000$ images of the set were split into $45,000$ for training, $5,000$ for validation, and $10,000$ for the test set. Data pre-processing was done in the same way as~\cite{hidden-networks}. Models are trained during 200 epochs, using stochastic gradient descent (SGD) with weight decay $0.0005$ and momentum $0.9$. After a warmup of 5 epochs,  the learning rate is reduced using cosine annealing starting from $0.1$. The batch size is $128$ for all models except for the original hidden-networks (HNNs)~\cite{hidden-networks}, for which the value used is $256$. Experiments using ImageNet use the hyperparameters recommended in~\cite{nvidia_hyperparam} for 100 epochs. Reported CIFAR100 accuracies are top-1 test accuracy scores of the models with the highest validation score, while for ImageNet they correspond to top-1 validation accuracy.
    
    \begin{table}[h!]
        \centering
        \vspace{-0.05cm}
        \begin{tabular}{|c|c|c|} 
         \hline
         Method & Architecture & Training \\
         \hline
         Standard (``Vanilla'')~\cite{resnet} & Feedforward & Weights\\
         Folding~\cite{liao_poggio} & \textbf{Recurrent} & Weights\\
         Hidden-Networks (HNN)~\cite{hidden-networks} & Feedforward & \textbf{Supermasks}\\
         \textbf{Hidden-Fold Networks (HFN)} & \textbf{Recurrent} & \textbf{Supermasks}\\
         \hline
        \end{tabular}
        \vspace{0.3cm}
        \caption{Summary of the four methods compared on ResNet in this paper.}
        \label{table:methods}
        \vspace{-0.3cm}
    \end{table}
    
    \vspace{-0.3cm}
    \subsection{Compatibility of supermasks and folding} \label{subsec:experiment_1}
    \vspace{-0.1cm}
    The results of applying both supermasks and folding suddenly to the whole network would be hard to interpret. Since starting to fold from the first stages could potentially destroy the feature hierarchy at its base, and there is more evidence for the fourth stage of the visual ventral stream to have a recurrent structure~\cite{kar2019evidence}, the first set of experiments fold progressively more stages of ResNet50 starting from the last one. Additionally, different combinations of weight initialization (Kaiming normal and SC) and BatchNorm (with and without UBN) are tested. These models were trained on CIFAR100 using dense learned weights (\figurename~\ref{fig:incremental_folding}) and using supermasks with a fixed density value of top-$k$\%$=50\%$ (\figurename~\ref{fig:incremental_HFN}). 
    
    The results show a bigger loss in accuracy as more stages are folded, which is recuperated with UBN as expected. In the case of supermask training, both SC initialization and UBN boost the accuracy when used separately, and their effect is stronger when combined. Despite having fewer parameters, SC-initialized HFNs with UBN have similar accuracies to the folded models trained with dense learned weights and higher accuracy than the non-folded HNN-ResNet50. 
    SC and UBN are used for all HFN hereafter.
    
	\begin{figure}
	    \vspace{-0.1cm}
        \centering
        \begin{subfigure}[t]{0.38\textwidth}
            \includegraphics[width=\textwidth]{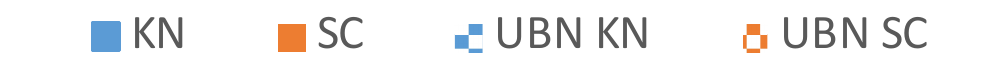}
            \label{fig:incremental_legend}
        \end{subfigure}\vspace{-0.5cm}
        \begin{multicols}{2}
            \centering
            \begin{subfigure}[t]{0.5\textwidth}
                \vspace{-0.7cm}
                \includegraphics[width=\textwidth]{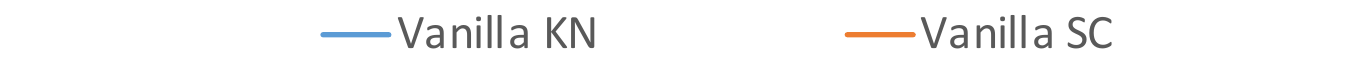}
                \label{fig:incremental_legend}
            \end{subfigure} \par
            \centering
            \begin{subfigure}[b]{0.5\textwidth}
                \vspace{-0.5cm}
                \includegraphics[width=\textwidth]{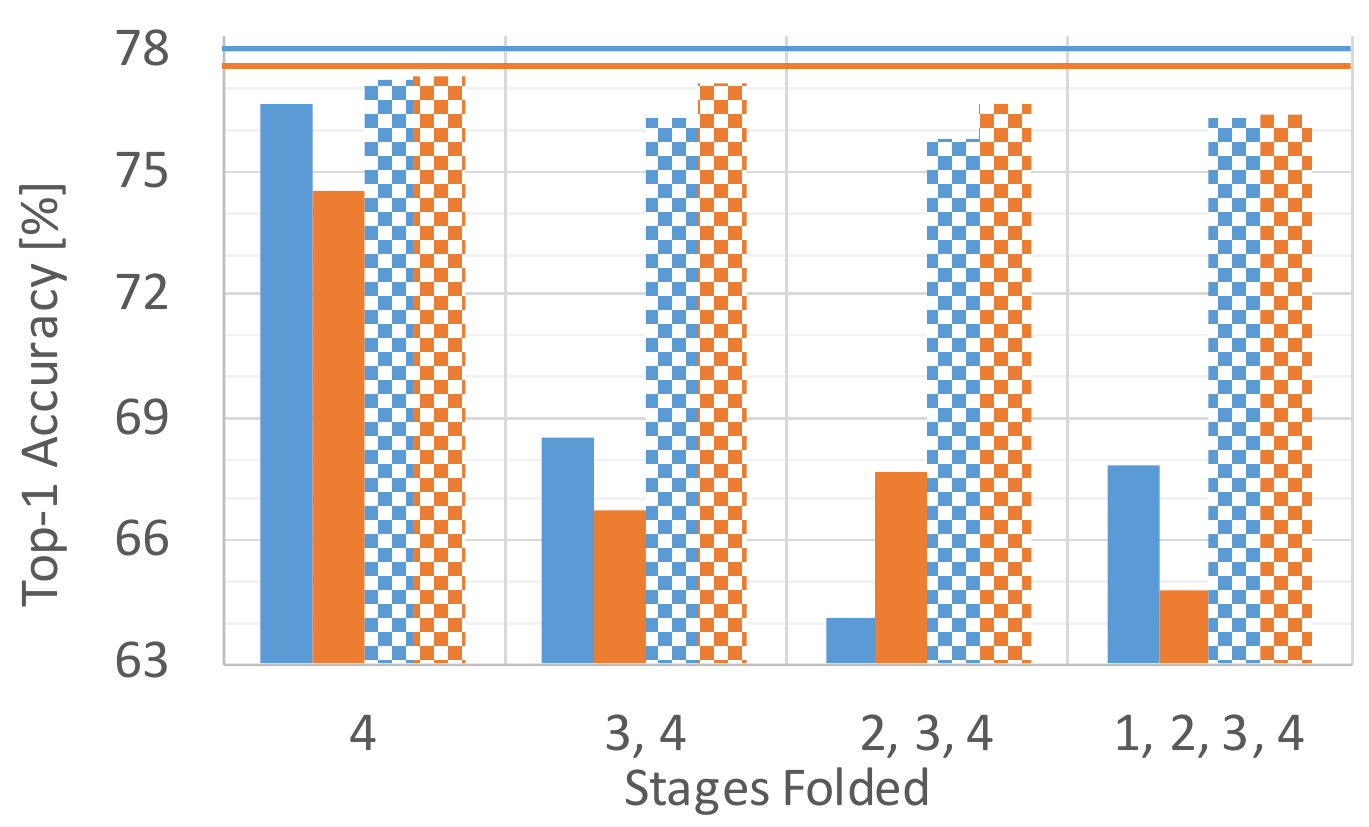}
                \vspace{-0.6cm}
                \caption{Weight training}
                \label{fig:incremental_folding}
            \end{subfigure}
            \centering
            \begin{subfigure}[t]{0.5\textwidth}
                \vspace{-0.7cm}
                \includegraphics[width=\textwidth]{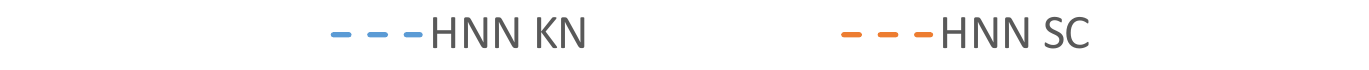}
                \label{fig:incremental_legend}
            \end{subfigure}\par
            \centering
            \begin{subfigure}[b]{0.5\textwidth}
                \vspace{-0.5cm}
                \includegraphics[width=\textwidth]{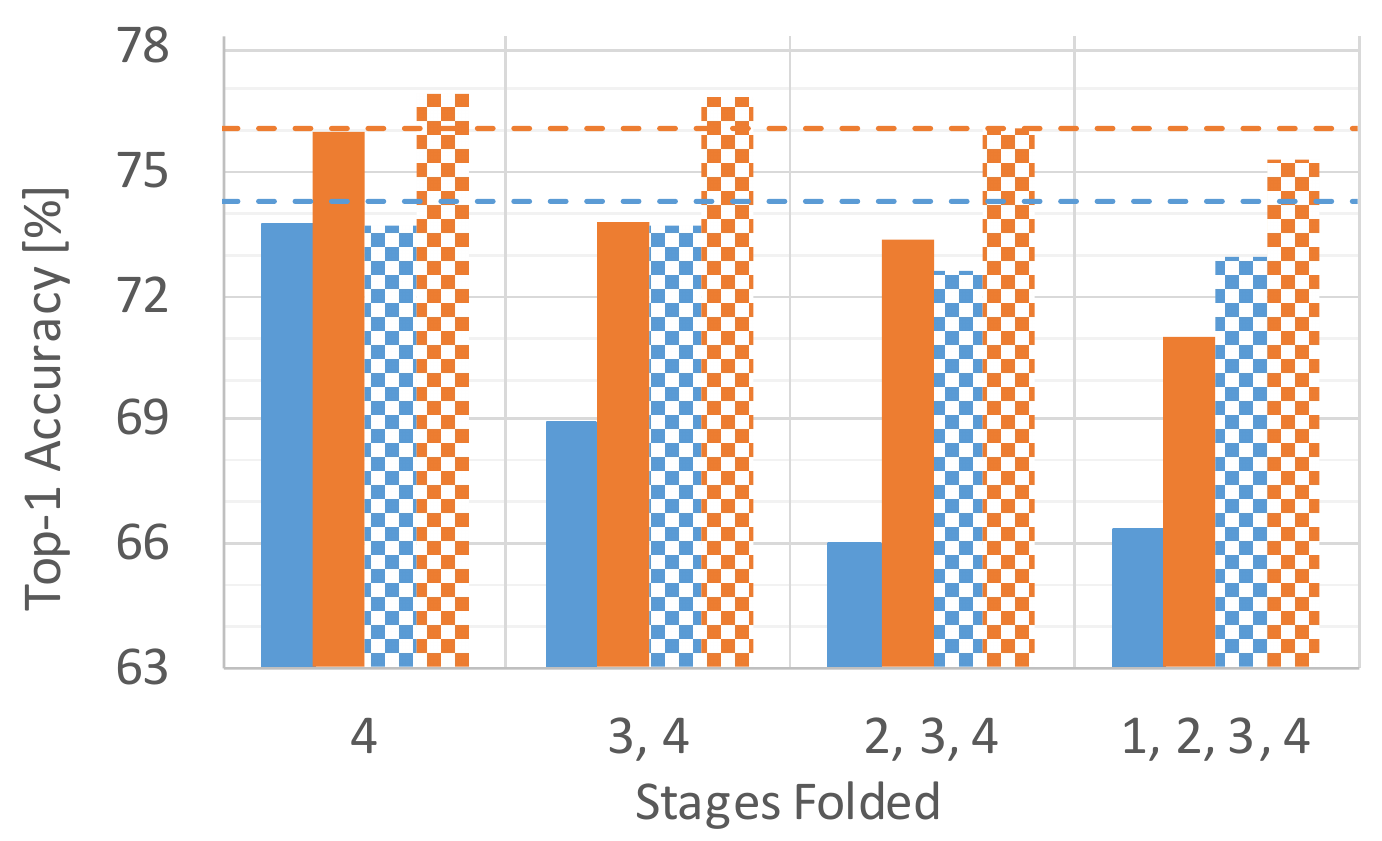}
                \vspace{-0.6cm}
                \caption{Supermask training}
                \label{fig:incremental_HFN}
            \end{subfigure}
        \end{multicols}
        \vspace{-0.2cm}
        \caption[caption]{ResNet50 and its hidden and/or folded variants trained on CIFAR100. KN: Kaiming normal; SC: signed Kaiming constant.
        (a) Vanilla and folding: standard training with weight learning.
        (b) HNN and HFN: training with supermasks (top-$k$\%$=50\%$).}
        \label{fig:foldin_n_arch}
        \vspace{-0.5cm}
    \end{figure}

\vspace{-0.3cm}
\subsection{Tuning supermask density} \label{subsec:experiment_2}
\vspace{-0.1cm}
    To check the tradeoffs between accuracy, supermask density, and amount of parameters, the following experiments test a range of top-$k$\% values for different configurations of folded stages. 
    \figurename~\ref{fig:topk_sweep_layers} shows that the optimal top-$k$\% values are between $20\%$ and $40\%$ for all cases and that folding only the last two stages (3 and 4) gives the highest accuracy.
    
	These ResNet50 configurations are compared in \figurename~\ref{fig:topk_sweep_param} with HNN, folding, and vanilla. The different points for each method correspond to the different top-$k$\% values in \figurename~\ref{fig:topk_sweep_layers}. \figurename~\ref{fig:topk_sweep_param} shows that HFN achieves equivalent or higher accuracy than folding or HNN with fewer parameters. Two conclusions can be drawn from these results: supermask training is as effective as weight learning for recurrent versions of ResNet, and folding ResNet improves the quality of the hidden subnetworks.
    
    Most ResNet weights are located in stages 3 and 4 due to the bigger feature map sizes and their higher number of blocks. Although folding more stages results in smaller models, reducing the supermask density provides higher compression gains with a smaller impact on accuracy. 
    Subsequent experiments only fold stages 3 and 4 since it provides the highest accuracy and a low number of parameters. Additionally, since optimizing top-$k$\% between $20\%$ and $40\%$ has a small impact on accuracy, all subsequent experiments use top-$k$\%$=30\%$.
     
    \afterpage{
	    \begin{figure}[t]
	        \vspace{-0.1cm}
            \centering
            \captionsetup[subfigure]{justification=centering}
            \begin{subfigure}[b]{0.5\textwidth}
                \centering
                \includegraphics[width=\textwidth]{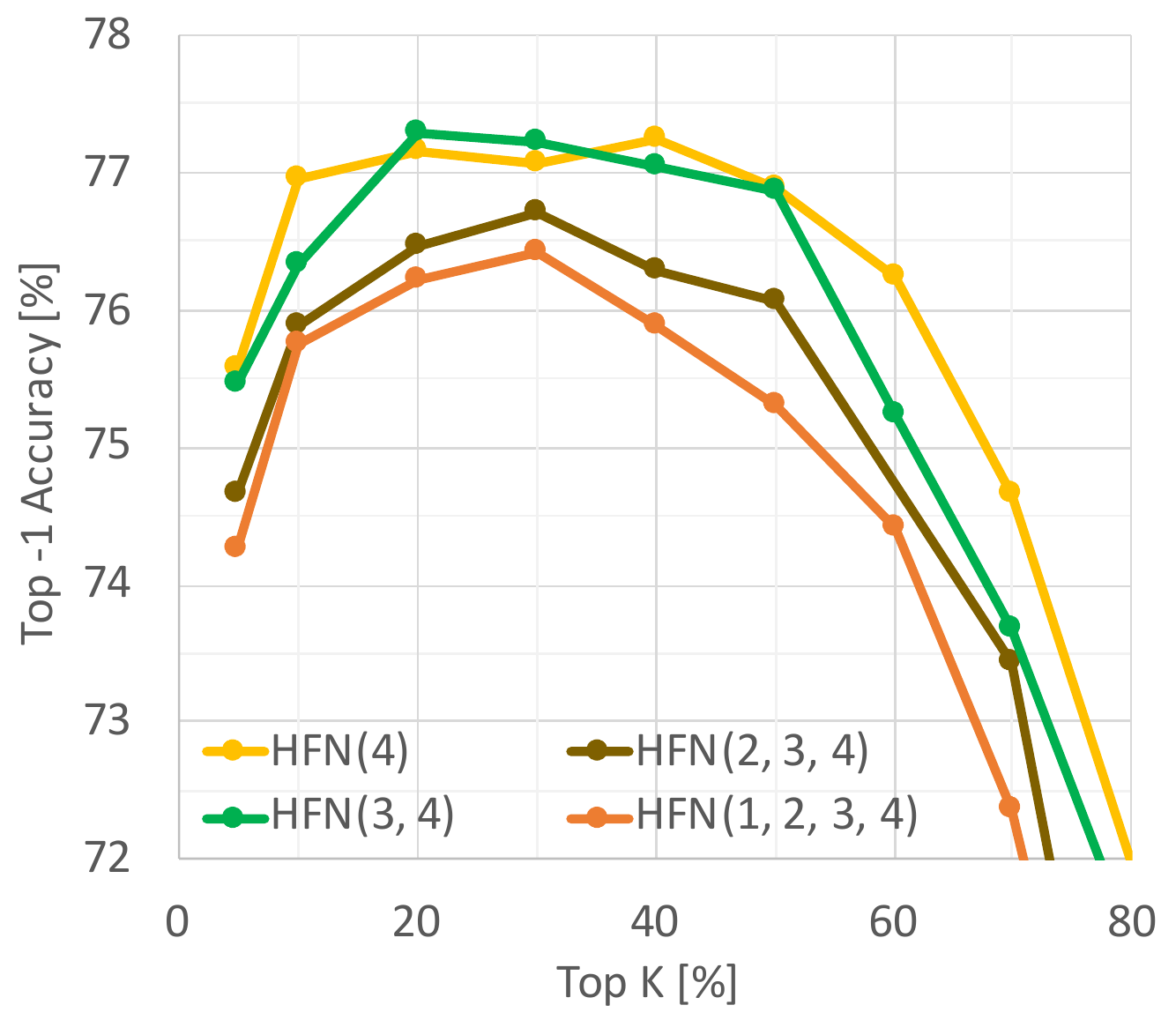}
                \vspace{-0.6cm}
                \caption{Accuracy to density tradeoff \\ among HFN configurations}
                \label{fig:topk_sweep_layers}
            \end{subfigure}\hfill
            \begin{subfigure}[b]{0.5\textwidth}
                \centering
                \includegraphics[width=\textwidth]{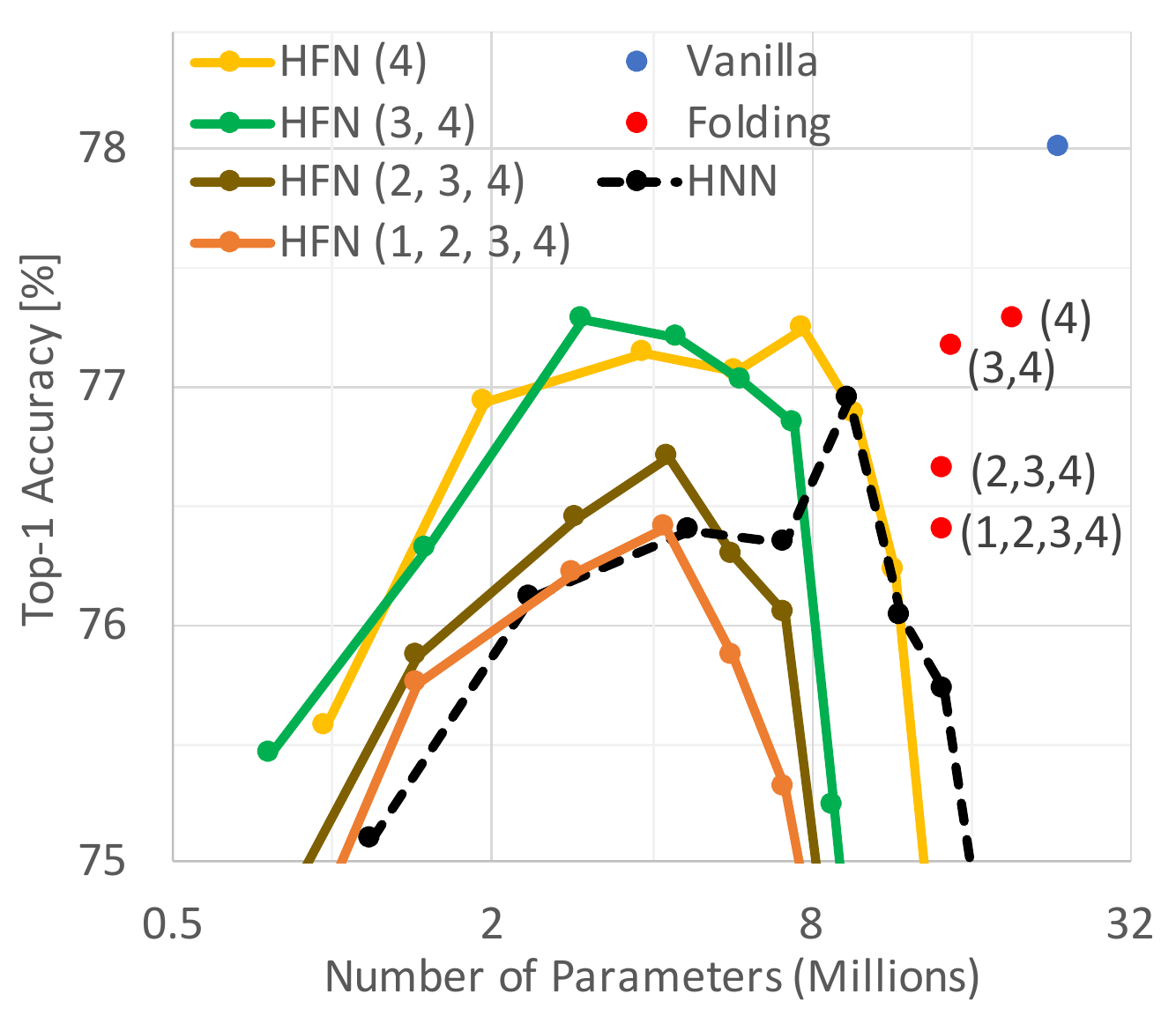}
                \vspace{-0.6cm}
                \caption{Accuracy to number of parameters tradeoff among the four tested methods}
                \label{fig:topk_sweep_param}
            \end{subfigure}
            \vspace{-0.05cm}
            \caption{Top-1 Accuracy results on CIFAR100 for ResNet50 trained with supermasks of different density value (top-$k$\%). Numbers in parenthesis indicate folded stages.}
            \label{fig:top_sweep}
            \vspace{-0.5cm}
        \end{figure}
    }
    
\vspace{-0.3cm}	
\subsection{Comparison of different model depths} \label{subsec:experiment_3}	
\vspace{-0.1cm}
    HFN's potential is seen more clearly when comparing ResNets of different sizes. Reference~\cite{hidden-networks} showed that a ResNet trained with supermasks has fewer parameters and equivalent accuracy to a deeper vanilla ResNet. Furthermore, a key advantage of folding ResNets into RNNs is that adding layers becomes equivalent to adding iterations, improving performance with few extra parameters, i.e., only those corresponding to additional UBN iterations.
    
    \figurename~\ref{fig:CIFAR100_param} compares ResNets of different depths and widths trained with the four discussed methods (vanilla, folding, HNN, and HFN). Models with more than 50 layers were trained for an additional 100 epochs, and all supermasks use top-$k$\%$=30\%$. Remarkably, HFNs prove to have the highest accuracies while also being the smallest models. An HFN version of ResNet152 has similar accuracy to ResNet50 while requiring $79.4\%$ fewer parameters, and an HFN-ResNet200 has a higher accuracy with $73.8\%$ fewer parameters.
    
    It is also worth noting that HFN accuracy grows monotonically with the number of layers. This suggests that additional iterations do improve accuracy and support the hypothesis that ResNets approximate unrolled RNNs. Nonetheless, there is also a difference in the number of non-folded blocks between these models. Future work should perform ablation studies to investigate this phenomenon more precisely. 

\vspace{-0.3cm}
\subsection{ImageNet experiments} \label{subsec:experiment_4}	
\begin{figure}[t]
        \centering
        \vspace{-0.1cm}
        \begin{subfigure}[b]{0.5\textwidth}
            \centering
            \includegraphics[width=\textwidth]{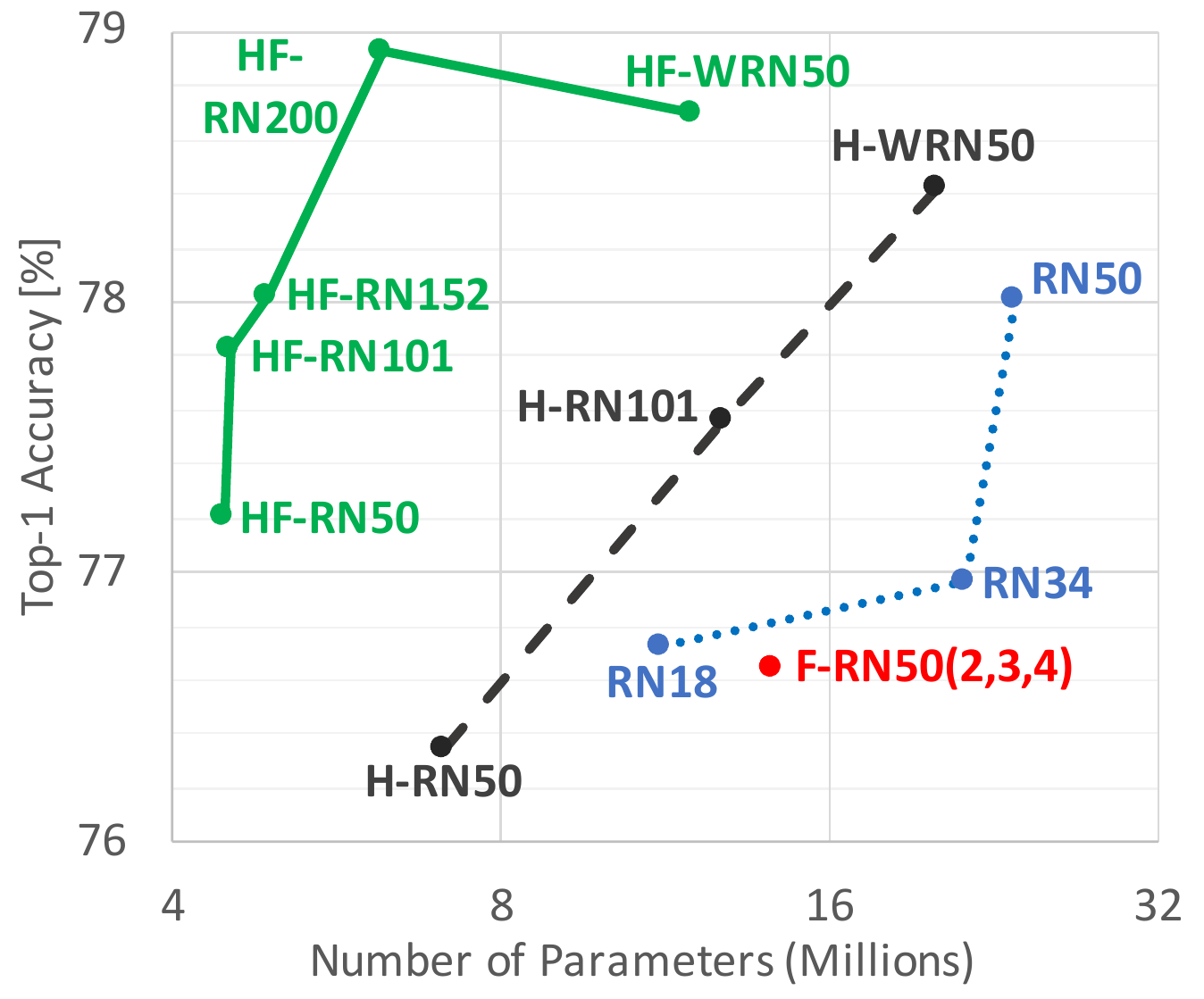}
            \vspace{-0.6cm}
            \caption{CIFAR100}
            \label{fig:CIFAR100_param}
        \end{subfigure}\hfill
        \begin{subfigure}[b]{0.5\textwidth}
            \centering
           \includegraphics[width=\textwidth]{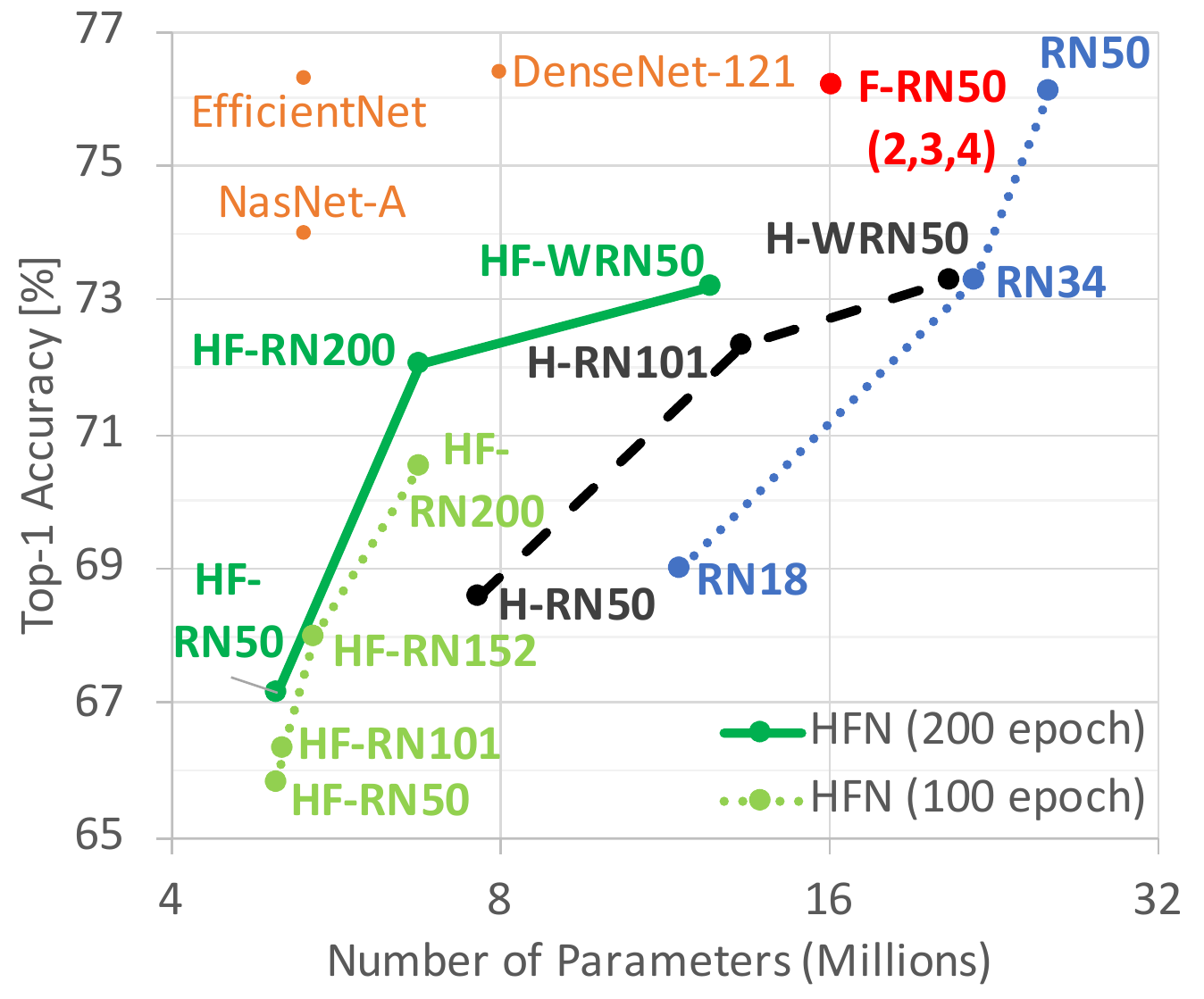}
            \vspace{-0.6cm}
            \caption{ImageNet}
            \label{fig:ImageNet_param}
        \end{subfigure}
        \begin{subfigure}[b]{0.5\textwidth}
            \centering
            \includegraphics[width=\textwidth]{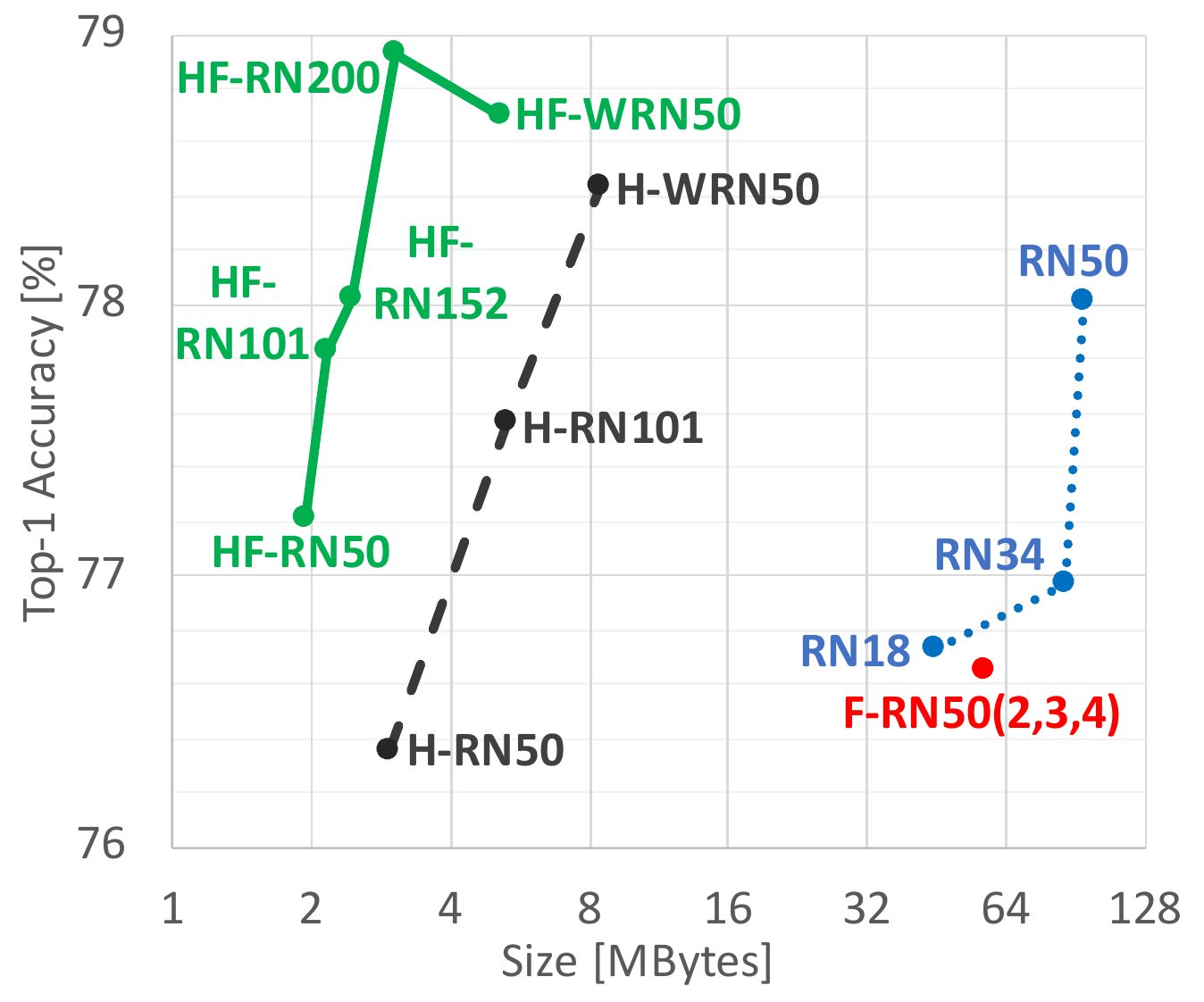}
            \vspace{-0.6cm}
            \caption{CIFAR100}
            \label{fig:CIFAR100_size}
        \end{subfigure}\hfill
        \begin{subfigure}[b]{0.5\textwidth}
            \centering
            \includegraphics[width=\textwidth]{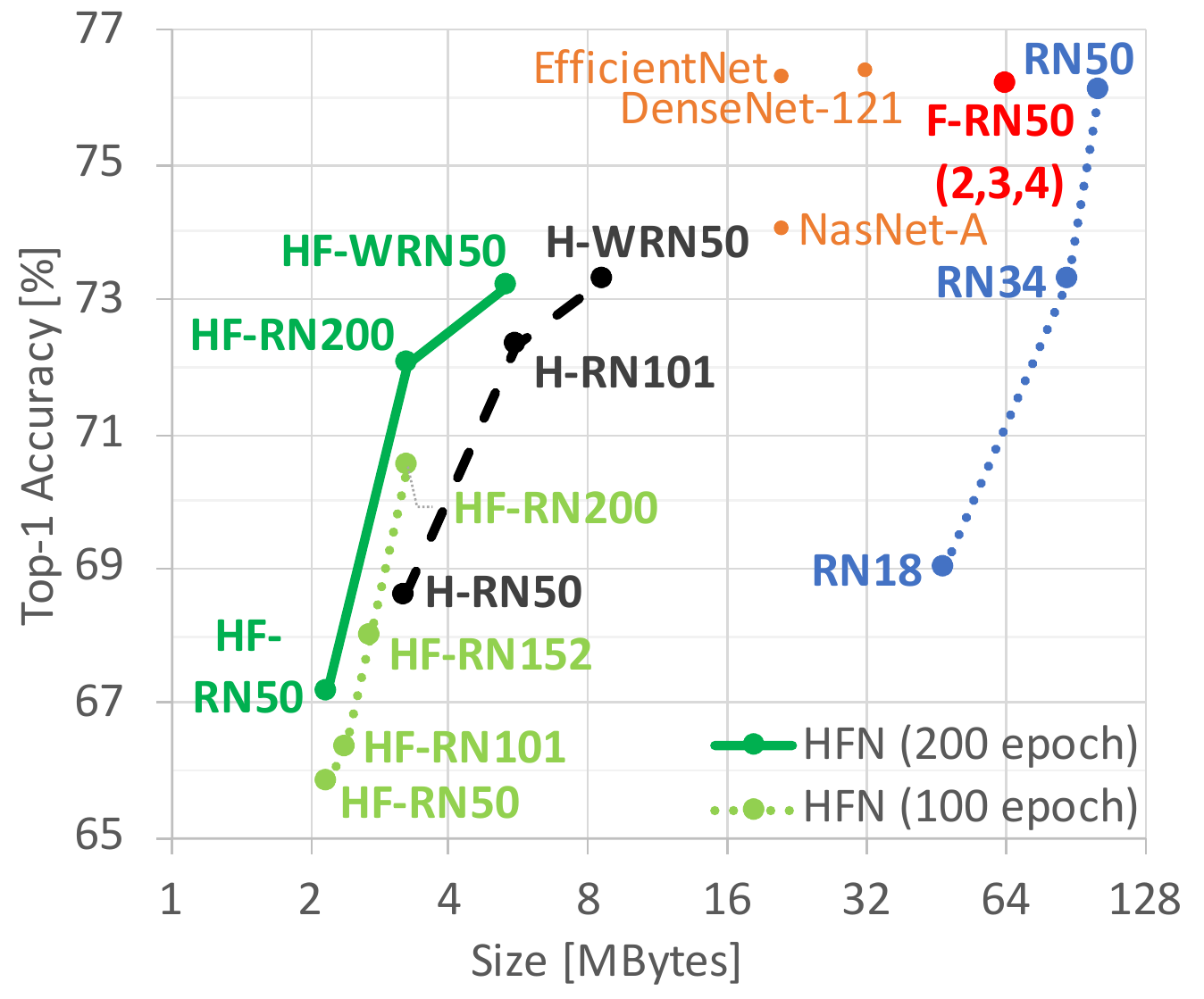}
            \vspace{-0.6cm}
            \caption{ImageNet}
            \label{fig:ImageNet_size}
        \end{subfigure}
        \vspace{-0.1cm}
        \caption[caption]{
        Top-1 Accuracy of ResNets of different depth and width. Architectures are abbreviated as RN: ResNet; WRN: Wide ResNet. Methods are indicated as F: Folded; H: HNN; HF: HFN; otherwise, vanilla. Numbers in parenthesis indicate folded stages (3 and 4 for all HFN). Models with supermask use top-$k$\%$=30\%$.  (a) and (b) compare number of parameters, while (c) and (d) compare model memory size. Non-ResNet models in orange.}
        \label{fig:params_n_sizes}
        \vspace{-0.6cm}
    \end{figure}

\vspace{-0.1cm}
    \figurename~\ref{fig:ImageNet_param} shows the model comparison in Section \ref{subsec:experiment_3}, now using the ImageNet dataset. While the results are less impressive than with CIFAR100, HFN still achieves competitive accuracy on ImageNet. HFN-ResNet200 shows similar accuracy to HNN-ResNet101 or ResNet34, with $49.2\%$ and $68.9\%$ less parameters, respectively. It is also significantly more accurate than HNN-ResNet50, which has a similar number of parameters. Even though vanilla ResNet50 achieved better accuracy, its number of parameters is much larger.
    
    Future research on optimal training schedules for supermask training should help HFN to achieve even better results.
    HFN's learning converges in all CIFAR100 cases, but in the case of ImageNet the loss kept slowly dropping without reaching convergence in a reasonable amount of epochs.
    RAdam~\cite{radam} achieved slightly better results than SGD in some cases, and extending training by 100 epochs consistently produced more accurate HFNs.

\vspace{-0.3cm}
\subsection{Model memory size} \label{subsec:experiment_5}
\vspace{-0.1cm}
   All the graphs discussed above compare accuracy based on the number of parameters. Since all the discussed methods use a numeric precision of 32-bit, these graphs can be translated directly into comparisons of the memory required to store each model. However, if considering specialized hardware for inference, a more insightful comparison can be made by exploiting the compression potential of supermasks and SC initialization.
   
   Since the weights are never updated, it is only necessary to store the supermask and the seed for generating the random signs, with the $\sigma$ values calculated on runtime from the network's shape. Additionally, the supermask only needs one bit per weight to indicate whether if the corresponding connection is part of the subnetwork or not. This straightforward compression vastly reduces the memory needed to store a trained model. Although it is not performed in this paper, supermasks could be compressed even further by exploiting their low density with sparse coding, similarly to~\cite{deepcomp}. In the case of HFN, it is necessary to store the scales and biases of UBN. However, the reduction in weights gained from folding far outweighs the cost of the additional UBN parameters, as it can be appreciated in \figurename~\ref{fig:CIFAR100_size} for CIFAR100, and in \figurename~\ref{fig:ImageNet_size} for ImageNet. \figurename~\ref{fig:params_n_sizes} also compares HFN to other small-sized models (EfficientNet\cite{efficientnet}, DenseNet\cite{densenet}, and NasNet\cite{nasnet}), and shows that although HFN has a lower parameter efficiency, it offers much larger memory gains.
   
   \tableautorefname~\ref{table:same_model} shows the effects of the different discussed methods on the same model (ResNet50). Although there is a moderate drop in accuracy, especially in the case of ImageNet, HFN-ResNet50 reduces weight memory storage to a mere $1.95$MB on CIFAR100 and $2.18$MB on ImageNet, small enough to fit in on-chip memory. More importantly, these size gains hold in comparisons of models of similar accuracy, summarized in \tableautorefname~\ref{table:same_acc}. Remarkably, HFN-ResNet152 has similar accuracy to ResNet50 on CIFAR100 despite being $38.5$x smaller, and on ImageNet HFN-ResNet200 has similar accuracy to ResNet34 while being $26.8$x smaller.
   
    \begin{table}[h!]
    \centering
    \vspace{-0.05cm}
    \begin{tabular}{|c|c|r|r|r|r|} 
     \hline
     Dataset & Model 
     & \begin{tabular}{@{}c@{}}Top-1 \\  {Acc. [\%]}\end{tabular} 
     & \begin{tabular}{@{}c@{}}Parameters \\ {(Millions)}\end{tabular} 
     & \begin{tabular}{@{}c@{}}Size \\ {[MB]}\end{tabular} 
     & \begin{tabular}{@{}c@{}}Size \\ Reduction\end{tabular} \\
     \hline
     CIFAR100   & ResNet50                      & 78.01          & 23.71    & 94.82             & - \\
                & Folded ResNet50 (2,3,4)       & 76.65          & 14.24    & 56.94              & 1.67x \\
                & HNN-ResNet50                  & 76.35          & 7.10     & 3.00              & 31.66x \\
                & \textbf{HFN-ResNet50}         & 77.21          & 4.45     & \textbf{1.95}     & \textbf{48.71x} \\
     \hline
     ImageNet   & ResNet50                          & 76.10          & 25.55    & 102.22             & - \\
                & Folded ResNet50 (2,3,4)           & 76.20          & 16.08    & 64.34              & 1.59x \\
                & HNN-ResNet50                      & 68.60          & 7.65     & 3.19              & 32.04x \\
                & \textbf{HFN-ResNet50}             & 67.70          & 5.00     & \textbf{2.18}     & \textbf{46.89x} \\
     \hline
    \end{tabular}
    \vspace{0.3cm}
    \caption{Accuracy and size effects of the different methods, compared on the same model.}
    \label{table:same_model}
    \vspace{-0.3cm}
    \end{table}
   
    \begin{table}[h!]
    \centering
    \vspace{-0.2cm}
    \begin{tabular}{|c|c|r|r|r|r|} 
     \hline
     Dataset & Model 
     & \begin{tabular}{@{}c@{}}Top-1 \\  {Acc. [\%]}\end{tabular} 
     & \begin{tabular}{@{}c@{}}Parameters \\ {(Millions)}\end{tabular} 
     & \begin{tabular}{@{}c@{}}Size \\ {[MB]}\end{tabular} 
     & \begin{tabular}{@{}c@{}}Size \\ Reduction\end{tabular} \\
     \hline
     CIFAR100   & ResNet50                  & 78.01     & 23.71     & 94.82             & - \\
                & HNN-WideResNet50          & 78.43     & 20.08     & 8.37              & 11.32x \\
                & \textbf{HFN-ResNet200}    & 78.93     & 6.21      & \textbf{3.02}     & \textbf{31.40x} \\
                & \textbf{HFN-ResNet152}    & 78.02     & 4.88      & \textbf{2.46}     & \textbf{38.54x} \\
     \hline
     ImageNet   & ResNet34                  & 73.30     & 21.78     & 87.19             & - \\ 
                & HNN-WideResNet50          & 73.30     & 20.64     & 8.60              & 10.14x \\
                & \textbf{HFN-WideResNet50} & 73.19     & 12.50     & \textbf{5.34}     & \textbf{16.33x} \\
                & \textbf{HFN-ResNet200}    & 72.06     & 6.77      & \textbf{3.25}     & \textbf{26.83x} \\
     \hline
    \end{tabular}
    \vspace{0.3cm}
    \caption{Size comparison of models of similar accuracy, using the proposed compression.}
    \label{table:same_acc}
    \vspace{-0.3cm}
    \end{table}
    
    \vspace{-0.3cm}
\section{Discussion} \label{sec:discussion}
\begin{figure}
    \vspace{-0.1cm}
    \centering
    \begin{subfigure}[t]{0.6\textwidth}
            \hspace{1cm}\includegraphics[width=\textwidth]{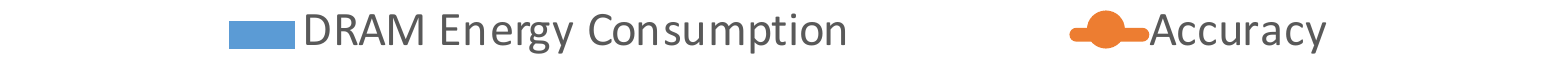}
        \label{fig:incremental_legend}
    \end{subfigure}\vspace{-0.4cm}
    \begin{multicols}{2}
        \centering
        \begin{subfigure}[b]{0.5\textwidth}
            \vspace{-0.5cm}
            \includegraphics[width=\textwidth]{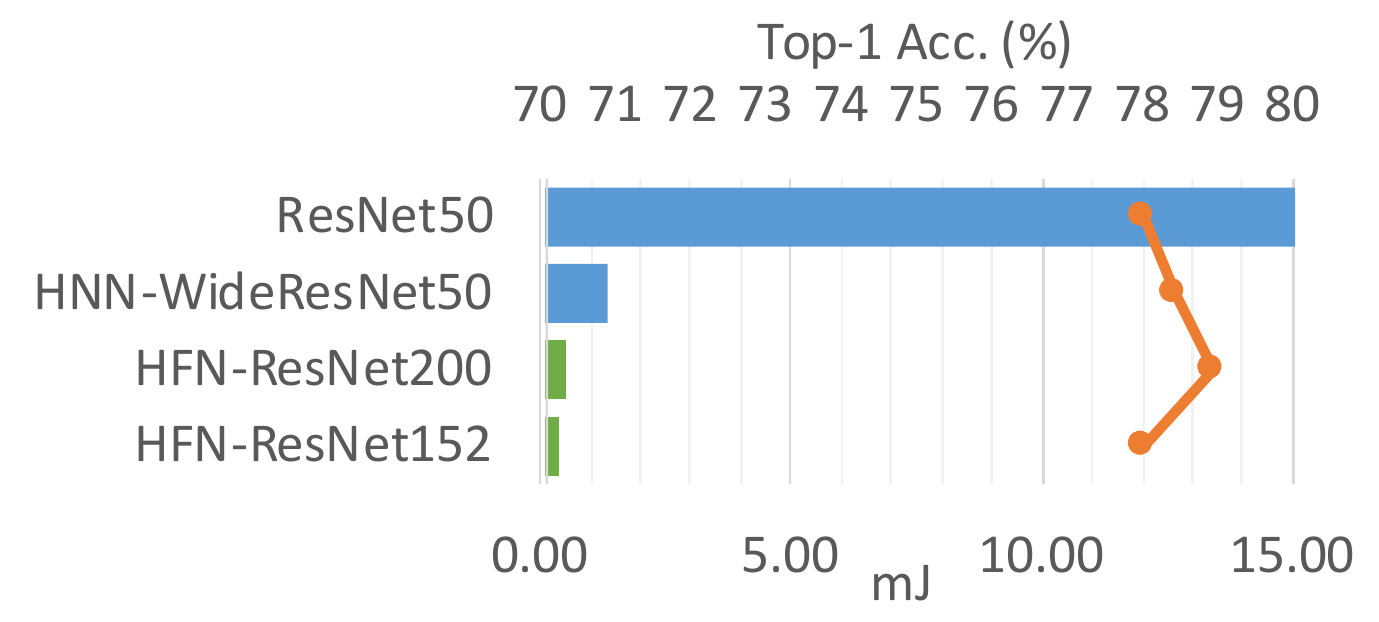}
            \vspace{-0.6cm}
            \caption{CIFAR100}
            \label{fig:incremental_folding}
        \end{subfigure}
        \centering
        \begin{subfigure}[b]{0.5\textwidth}
            \vspace{-0.5cm}
            \includegraphics[width=\textwidth]{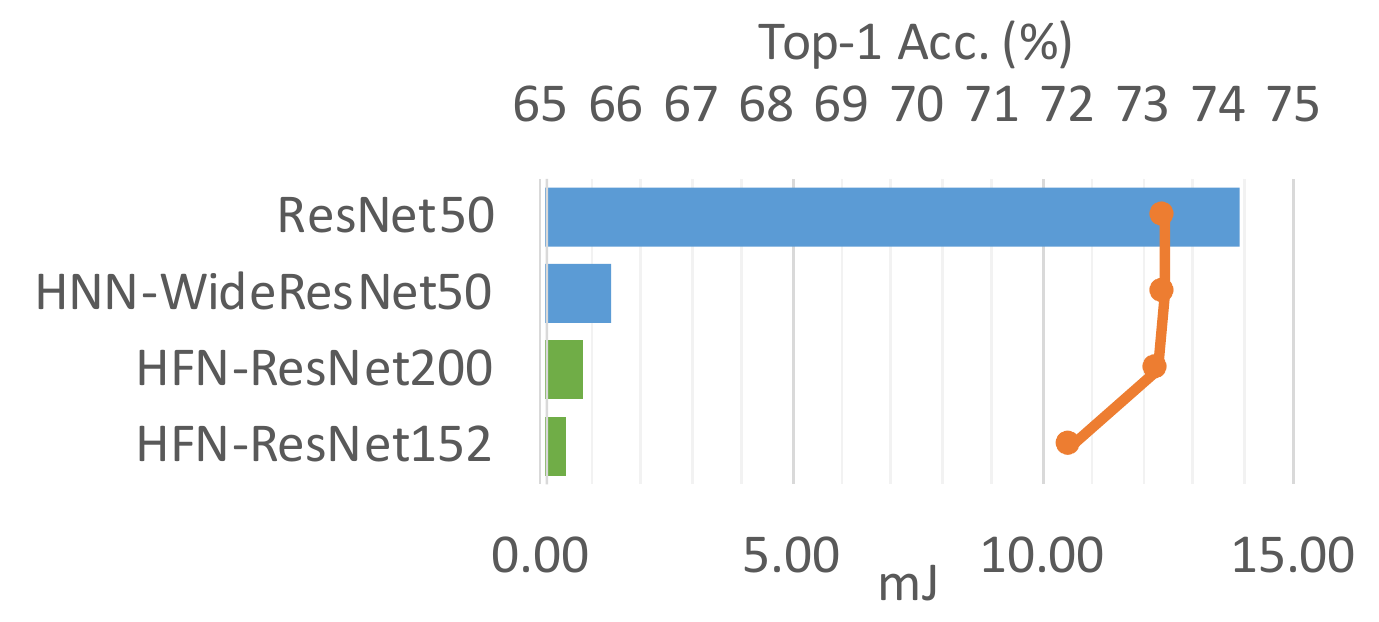}
            \vspace{-0.6cm}
            \caption{ImageNet}
            \label{fig:incremental_HFN}
        \end{subfigure}
    \end{multicols}
    \vspace{-0.2cm}
    \caption[caption]{Estimation of energy consumed for loading models of similar accuracy from DRAM, considering an ideal $45$~nm CMOS inference accelerator.}
    \label{fig:energy}
    \vspace{-0.5cm}
\end{figure}

\vspace{-0.1cm}
	HFNs are accurate models with few and highly reused random parameters. When considering specialized hardware inference accelerators with a random number generator and the capability of operating with binary supermasks, HFNs can be compressed enough to be stored on on-chip SRAM, making off-chip DRAM access for parameters unnecessary. This promises vast energy-efficiency advantages.
	The bulk of computations used for neural networks consists of energy-hungry multiplications. Still, this cost is minute compared to that of off-chip memory access. In the case of $45$~nm CMOS, a 32 bit floating-point multiplication consumes $3.7$~pJ, while a 32 bit DRAM read costs $640$~pJ~\cite{horowitz20141,han2016eie}. Specialized hardware often operates with reduced numeric precision, making the contrast even starker: a 16 bit floating-point multiplication uses $291$x less energy than DRAM access. Additionally, off-chip memory also suffers from a much longer latency. With random weights, highly compressed binary supermasks, and recurrent connections reusing parameters, memory reads can be reduced to a minimum. \figurename~\ref{fig:energy} shows a comparison of the estimated energy consumed by DRAM access for loading models of similar accuracy, considering an ideal accelerator implemented in $45$~nm CMOS. HFN reduces energy consumption by two orders of magnitude, making it a promising candidate for implementing energy-efficient DNN acceleration hardware.

    On the other hand, it shall be noted that HFN is not aimed at CPU/GPU implementation. Since the parameters must be uncompressed at runtime and folding does not change the number or size of convolutions, HFN-ResNet50's computational cost on standard processors is similar to ResNet50's. HFN-ResNet152 has similar accuracy to ResNet-50 while using 4.9x less parameters (\tableautorefname~\ref{table:same_acc}, CIFAR100), but its computational cost is almost 3x higher. This tradeoff between computation and size is avoided on specialized hardware, as the computational cost is also reduced by exploiting the model's sparsity and processing supermasks with binary operations. For software implementations, HFN could be coupled with techniques for reducing computational cost, such as reduced numerical precision, depth-wise separable convolutions~\cite{mobilenets}, or AdderNets~\cite{addernet}, or by generalizing folding to be compatible with the dense connections of DenseNet~\cite{densenet}, which would reduce HFN's number of channels.
    
    Several aspects of supermask training still have to be addressed. Future work should find a method for determining the optimal supermask density a priori or optimizing it during the training process. Furthermore, tuning different density values for each layer could reduce size while increasing accuracy. Alternatively, recent work has proposed using a global density value instead of constraining the density of each layer~\cite{probmask}. An initialization method that eliminates the need for normalization, similar to Fixup~\cite{fixup} but compatible with supermask training, could potentially remove the additional parameters introduced by UBN. Even though HFN achieves remarkable model size reduction, supermask training introduces an additional training cost by using a separate score tensor, which should be addressed. Training ResNet50 on CIFAR100 using 1xNVIDIA GeForce RTX 3090 takes $44$~s per epoch, whereas its HNN and HFN versions take $153$~s and $149$~s respectively.
    
    \vspace{-0.4cm}
\section{Conclusion} \label{sec:conclusion}
\vspace{-0.1cm}
    Hidden-fold networks combine the advances of two recent research trends into a residual network that is small yet accurate. 
    When using an optimal supermask density and training schedule, this method proves beneficial for both original trends: 
    folded models are more accurate when trained with supermasks, and folding ResNets yields more accurate hidden subnetworks than strictly feed-forward models.
    HFN's random weights do not need to be stored, as they are substituted with a random seed and sparse binary supermasks that can be highly compressed. Furthermore, recurrent connections can be exploited for reusing parameters. Consequently, HFN can be implemented in specialized hardware with a tiny number of off-chip memory accesses, guaranteeing an energy-efficient and faster system.
    \vspace{-0.4cm}
\section*{Acknowledgement}
\vspace{-0.3cm}
    This work was supported by JST CREST Grant Number JPMJCR18K2, Japan.
\vspace{-0.1cm}

    \bibliography{IEEEabrv, abrv, egbib}

\end{document}